\documentclass{amia}

\usepackage{float}

\usepackage[normalem]{ulem}
\useunder{\uline}{\ul}{}
\usepackage[table,xcdraw]{xcolor}

\usepackage[utf8]{inputenc} 
\usepackage[T1]{fontenc}    
\usepackage{hyperref}       
\usepackage{url}            
\usepackage{booktabs}       
\usepackage{amsfonts}       
\usepackage{nicefrac}       
\usepackage{microtype}      
\usepackage{lipsum}

 \hypersetup{
     colorlinks=true,
     linkcolor=blue,
     filecolor=blue,
     citecolor = blue,      
     urlcolor=blue,
     }

\usepackage{epstopdf}
\usepackage{graphicx}
\usepackage{graphics}
\usepackage{amsmath}
\usepackage{hyperref}
\usepackage[symbol]{footmisc}

\usepackage{multirow}
\usepackage{array}
\usepackage{ltablex}
\usepackage[labelfont=bf]{caption}
\usepackage{colortbl}
\usepackage[superscript,nomove]{cite}
\usepackage{color}
\newcolumntype{C}{ >{\centering\arraybackslash} m{1.5cm} }

\usepackage[super]{natbib}
\usepackage{amssymb}     
\usepackage{graphicx}      
\usepackage{subcaption}   
\usepackage{xcolor}      
\usepackage[table]{xcolor}
\usepackage[margin=1in]{geometry}
\usepackage{wrapfig}

\usepackage{tikz}
\usetikzlibrary{shapes.geometric, arrows.meta, positioning, fit, backgrounds, calc, shadows.blur}
\usetikzlibrary{decorations.pathreplacing}

\definecolor{userbg}{HTML}{E8F0FE}
\definecolor{userborder}{HTML}{4285F4}
\definecolor{agentbg}{HTML}{FFF3E0}
\definecolor{agentborder}{HTML}{F09819}
\definecolor{pipebg}{HTML}{E8F5E9}
\definecolor{pipeborder}{HTML}{43A047}
\definecolor{outbg}{HTML}{FCE4EC}
\definecolor{outborder}{HTML}{E57373}
\definecolor{vizbg}{HTML}{F3E5F5}
\definecolor{vizborder}{HTML}{AB47BC}
\definecolor{chatbg}{HTML}{FFFDE7}
\definecolor{chatborder}{HTML}{FFB300}
\definecolor{geesebg}{HTML}{F1F8E9}
\definecolor{darktext}{HTML}{37474F}
\definecolor{chatmsg}{HTML}{FFF8E1}
\definecolor{chatmsgborder}{HTML}{FFD54F}

\usepackage{tikz}
\usetikzlibrary{positioning, arrows.meta, fit}

\definecolor{inputblue}{HTML}{E3F2FD}
\definecolor{inputborder}{HTML}{1976D2}
\definecolor{procgray}{HTML}{F5F5F5}
\definecolor{procborder}{HTML}{757575}
\definecolor{transformpurple}{HTML}{EDE7F6}
\definecolor{transformborder}{HTML}{5E35B1}
\definecolor{outputgreen}{HTML}{E8F5E9}
\definecolor{outputborder}{HTML}{43A047}
\definecolor{arrowgray}{HTML}{616161}

\usepackage{fancyhdr}
\fancypagestyle{firstpage}{
  \fancyhf{}

  \fancyfoot[L]{\vspace*{-8pt}\footnoterule\footnotesize
  Code: \url{https://github.com/DominoAI-Lab/GEESE-AMIA-2026} \\
  Demo: \url{https://huggingface.co/spaces/EvilRagdollCat/GEESE-HONK}}
}

\makeatletter
\providecommand{\@institutes}{}
\makeatother

\title{GEESE: Genotype-aware End-to-End Spatio-temporal Embedding for Behavioral Phenotyping}

\author{
Yiran Ding, MS$^1$; Yuen Gao, PhD$^2$; Chunqi Qian, PhD$^2$; Zijun Cui, PhD$^1$\\
$^1$Department of Computer Science and Engineering, Michigan State University, East Lansing, Michigan, USA\\
$^2$Department of Radiology, Michigan State University, East Lansing, Michigan, USA
}

\begin{document}

\maketitle
\thispagestyle{firstpage}

\section*{Abstract}

\textit{Behavioral phenotyping of genetic animal models currently requires labor-intensive manual feature engineering that limits reproducibility and scalability. We present GEESE, an end-to-end deep learning framework that learns behavioral representations directly from 3D pose dynamics without hand-crafted features. Using a pretrained time series foundation model, we encode movement sequences into a behavioral manifold that supports both behavior classification and genotype prediction. Evaluated across three autism-associated genetic models (CNTNAP2, CHD8, FMR1), our deep learning approach surpasses hand-crafted feature baselines in both tasks, revealing that learned representations capture genotype-specific behavioral signatures. The framework generalizes across genetic backgrounds, and an all-cohort model identifies both genetic background and genotype from movement patterns alone. We further provide HONK, an interactive intelligent tool enabling researchers without programming expertise to perform behavioral phenotyping from pose data through natural language interaction.}

\section{Introduction}

Behavioral assessment is fundamental to the diagnosis and treatment monitoring of neurological and psychiatric conditions, from motor symptoms in Parkinson's~\cite{jankovic2008parkinson} and Huntington's disease~\cite{bates2015huntington} to social and repetitive behaviors in Autism Spectrum Disorder (ASD)~\cite{hirota2023autism}. Despite this central role, clinical assessment remains largely subjective, time-intensive, and dependent on specialist availability. These limitations underscore the need for behavior phenotyping, which is the process of transforming raw movement into standardized, objective, and scalable digital signatures. Moving beyond qualitative observation, automated phenotyping would enable direct comparison of phenotypes across genetic models and accelerate drug development pipelines, while accessible screening could shorten the time from initial concern to diagnosis and intervention. 

While the objectives of behavioral phenotyping are clear in clinical settings, achieving them requires overcoming technical challenges. In preclinical research, rodent models carrying ASD-associated genetic variants are typically evaluated through behavior assays such as open field tests, social interaction paradigms, and grooming analysis~\cite{bey2014overview, kas2014assessing}. Modern markerless pose estimation methods have made high-resolution behavioral kinematic data increasingly accessible~\cite{mathis2018deeplabcut, pereira2019fast, pereira2022sleap, dunn2021geometric}. Recent advances in artificial intelligence, particularly deep learning and foundation models pretrained on large-scale data, offer an opportunity to automate this translation and reduce the barrier to clinical adoption of behavioral phenotyping tools. But translating raw pose coordinates into meaningful behavioral descriptions remains challenging, largely due to reliance on hand-crafted feature engineering that introduces researcher bias and limits reproducibility across laboratories. Existing deep learning approaches provide behavioral visualization and clustering but do not support genotype predictive downstream tasks.

In this work, we propose GEESE (Genotype-aware End-to-End Spatio-temporal Embeddings), a representation learning pipeline for behavioral phenotyping from 3D pose dynamics. We leverage a time series foundation model~\cite{guo2025foundation}, pretrained on large-scale temporal data as the encoder backbone, which is then fine-tuned on behavioral labels to learn a behavioral representations. The resulting latent space serves as a unified representation space supporting multiple downstream tasks without task-specific feature engineering.  
We evaluate GEESE on 146 recording sessions across three ASD-associated genetic models (CNTNAP2, CHD8, FMR1), showing that learned representations outperform hand-crafted baselines in both behavior classification and genotype prediction, generalize across cohorts, and capture genotype-specific behavioral signatures.

To support broader adoption, we additionally present HONK (Hands-On Natural-language Knowledgebase), an interactive analysis tool that enables researchers without programming expertise to perform behavioral phenotyping directly from pose data. This framework represents a step toward automated behavioral assessment tools that could ultimately support clinical screening and therapeutic development for ASD and other behaviorally-defined conditions.

\section{Related Works}

\noindent\textbf{Hand-crafted Feature Engineering for Behavioral Analysis.} Traditional behavioral analysis pipelines rely on hand-crafted feature engineering to manage the high dimensionality of kinematic data. Principal Component Analysis (PCA)~\cite{wold1987principal} is commonly used to reduce coordinates, while wavelet transforms~\cite{zhang2019wavelet, jia2022feature} approximate temporal dynamics. The s-DANNCE pipeline~\cite{klibaite2025mapping} combines 3D pose tracking with wavelet-based feature extraction to map behavioral structure, but still requires manually designed spectral features. These hand-crafted pipelines require extensive domain expertise to design appropriate features for each specific experimental context, inherently introducing researcher bias into the quantification process. Besides, the resulting metrics may miss subtle patterns not captured by predefined measurements. Additionally, subsecond behaviors are often missed~\cite{han2024multi}, and keypoint jitter can produce spurious segmentation transitions~\cite{weinreb2024keypoint}.

\noindent\textbf{Deep Learning Approaches for Behavioral Representation.} Learning behavioral representations directly from movement kinematics offers a fundamentally different paradigm that bypasses the limitations~\cite{schneider2023learnable} of human-defined features. Rather than summarizing motion through predefined descriptors, learning-based approaches embeds entire movement sequences into a continuous latent space~\cite{klibaite2025mapping}. Recent deep learning methods have been applied in this field: CEBRA uses contrastive learning to produce consistent latent embeddings from behavioral and neural data~\cite{schneider2023learnable}; Keypoint-MoSeq applies generative modeling to parse continuous pose trajectories into discrete behavioral syllables~\cite{weinreb2024keypoint}; the Social Behavior Atlas (SBeA) framework uses few-shot learning to reduce annotation requirements for pose estimation and identity recognition~\cite{han2024multi}. These approaches demonstrate that end-to-end representation learning can extract behavioral structure from movement data without hand-crafted features. However, none of these methods support genotype prediction from learned representations, leaving an end-to-end predictive framework absent from existing behavioral analysis pipelines.

\section{Methodology}

\begin{figure}[!htbp]
\centering
\includegraphics[width=\textwidth]{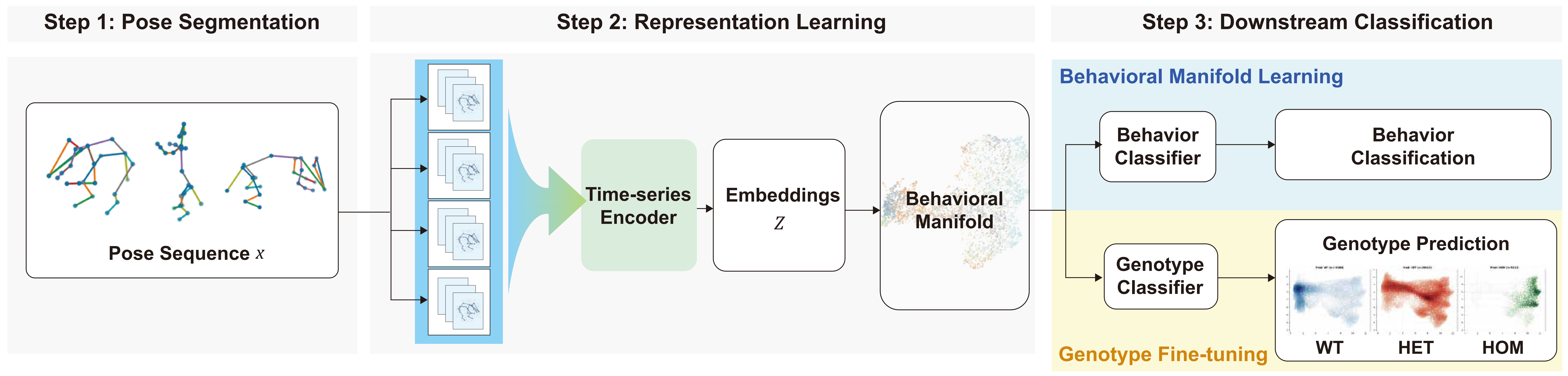}
\caption{System architecture. Pose sequences are processed by a pretrained time series model, producing behavioral representations. Training on behavioral labels organizes these representations by behavior type, enabling behavior classification. The same representations support genotype prediction after brief fine-tuning on genotype labels.}
\label{fig:geesepipeline}
\end{figure}

GEESE operates as a three-step pipeline (Figure~\ref{fig:geesepipeline}). First, continuous 3D pose recordings are segmented into overlapping temporal windows, each represented as a matrix $\mathbf{X} \in \mathbb{R}^{T \times D}$ capturing the instantaneous posture and short-term dynamics of all skeletal keypoints. Second, each window is passed through a pretrained time series foundation model that compresses the high-dimensional pose sequence into a compact embedding $\mathbf{z} \in \mathbb{R}^{d}$. The embeddings form a behavioral manifold where proximity reflects kinematic similarity. Task-specific classification heads are then attached for downstream prediction.

\subsection{Preliminary: Time Series Foundation Models}

MOMENT~\cite{goswami2024moment} is an open-source foundation model pretrained on the Time Series Pile, a large-scale collection spanning healthcare, finance, and engineering domains. Its transformer encoder ($L = 24$ layers) processes multivariate time series through patch embedding and supports transfer learning to classification tasks.

\subsection{Model Architecture}

The encoder $f_\theta$ maps each input window to a fixed-dimensional embedding:
\begin{equation}
    \mathbf{z} = f_\theta(\mathbf{X}) \in \mathbb{R}^{d}
\end{equation}
where $d = 1024$. Internally, MOMENT processes the $D = 69$ input channels through a patch embedding layer, applies $L = 24$ transformer encoder blocks with multi-head self-attention, and produces per-token representations that are aggregated via mean pooling:
\begin{equation}
    \mathbf{z} = \frac{1}{N_{\text{patches}}} \sum_{i=1}^{N_{\text{patches}}} \mathbf{h}_i
\end{equation}
where $\mathbf{h}_i \in \mathbb{R}^{d}$ is the output of the final transformer block for patch $i$, and $N_{\text{patches}}$ is the number of temporal patches. 

For downstream tasks, we attach task-specific linear classification heads to the encoder. For behavior classification, a linear head $g_{\phi_b}$ maps the embedding to class logits:
\begin{equation}
    \hat{\mathbf{y}}_b = g_{\phi_b}(\mathbf{z}) = \mathbf{W}_b \mathbf{z} + \mathbf{b}_b
\end{equation}
where $\mathbf{W}_b \in \mathbb{R}^{C_b \times d}$, $\mathbf{b}_b \in \mathbb{R}^{C_b}$, and $C_b = 9$ is the number of behavior categories. For genotype classification, a separate linear head $g_{\phi_g}$ maps to genotype logits:
\begin{equation}
    \hat{\mathbf{y}}_g = g_{\phi_g}(\mathbf{z}) = \mathbf{W}_g \mathbf{z} + \mathbf{b}_g
\end{equation}
where $\mathbf{W}_g \in \mathbb{R}^{C_g \times d}$, $\mathbf{b}_g \in \mathbb{R}^{C_g}$, and $C_g \in \{2, 3\}$ depending on the cohort. 

\subsection{Training Strategy}
\label{sec:training}
We adopt a two-stage training strategy: the encoder is first trained on expert-annotated behavioral labels to organize the manifold by behavior type, then fine-tuned end-to-end on genotype labels with a reduced learning rate to gain sensitivity to genotype-associated movement patterns while preserving learned behavioral structure. 

\noindent\textbf{Stage 1: Supervised Training for Behavior Classification.}
HLAC (High-Level Action Classification), manually defined by the developers of s-DANNCE based on visual inspection, provides human-annotated behavioral labels including locomotion, grooming, rearing, and other stereotyped actions~\cite{klibaite2025mapping}. We train the model by minimizing the cross-entropy loss over behavior labels:
\begin{equation}
    \mathcal{L}_{\text{behav}} = -\frac{1}{N} \sum_{i=1}^{N} \sum_{c=1}^{C_b} y_{b,i}^{(c)} \log \frac{\exp(\hat{y}_{b,i}^{(c)})}{\sum_{j=1}^{C_b} \exp(\hat{y}_{b,i}^{(j)})}
\end{equation}
where $y_{b,i}^{(c)}$ is the one-hot encoded ground truth label for sample $i$ and class $c$, and $N$ is the number of training samples. During this stage, the encoder parameters $\theta$ are frozen and only the classification head parameters $\phi_b$ are updated.

\noindent\textbf{Stage 2: Fine-tuning for Genotype Classification.}
We then replace the classification head and fine-tune on genotype labels. The genotype loss is:
\begin{equation}
    \mathcal{L}_{\text{geno}} = -\frac{1}{N} \sum_{i=1}^{N} \sum_{c=1}^{C_g} y_{g,i}^{(c)} \log \frac{\exp(\hat{y}_{g,i}^{(c)})}{\sum_{j=1}^{C_g} \exp(\hat{y}_{g,i}^{(j)})}
\end{equation}
During this stage, both the encoder parameters $\theta$ and the genotype head parameters $\phi_g$ are updated with a reduced learning rate ($\eta = 10^{-5}$) to preserve learned behavioral structure while gaining sensitivity to genotype-associated movement patterns. Training runs for up to 10 epochs with early stopping (patience 5).

\section{Evaluation}

\subsection{Experimental Settings}

\noindent\textbf{Datasets.}
We used 3D pose tracking data from the s-DANNCE repository ~\cite{klibaite2025mapping}, which contains recordings from rodent models of autism-associated genes. We analyzed three cohorts: CNTNAP2 (42 sessions; WT, HET, HOM), CHD8 (80 sessions; WT, HET), and FMR1 (24 sessions; WT, HET). All recordings were lone (single-animal) sessions. Each session contains continuous 3D coordinates for 23 skeletal keypoints captured at 30 fps, with expert-annotated behavioral labels for nine categories (idle, sniff, groom, scrunch, active crouch, rearing, explore, locomotion, fast locomotion). Behavioral annotations were defined by the s-DANNCE developers based on visual inspection of rodent pose dynamics. For data preprocessing, continuous pose sequences were segmented into overlapping windows for model input. Each window is represented as a matrix $\mathbf{X} \in \mathbb{R}^{T \times D}$, where $T = 32$ frames (approximately 1 second at 30 fps) is the window length and $D = J \times 3 = 69$ is the number of input channels obtained by flattening the 3D coordinates of $J = 23$ skeletal keypoints. Windows are extracted with stride $S = 16$ for 50\% overlap. To prevent data leakage from temporally correlated windows, we employed session-based splitting: train (64\%), validation (16\%), and test (20\%), ensuring all windows from a given session appear exclusively in one split.

\noindent\textbf{Evaluation Metrics.}
We report test accuracy and normalized confusion matrices for both behavior and genotype classification. For unsupervised analysis, we apply K-Means clustering (k=9) and evaluate using silhouette score~\cite{rousseeuw1987silhouettes} for cluster compactness and normalized mutual information (NMI)~\cite{strehl2002cluster} for correspondence with ground truth labels, in order to examine learned representations beyond supervised labels.

\noindent\textbf{Implementation Details.} Both stages use Adam with mixed-precision (FP16) training. Stage~1 uses learning rate reduction on plateau (factor 0.5, patience 5); Stage~2 uses a fixed learning rate of $10^{-5}$ with early stopping (patience 5).

\subsection{Quantitative Evaluation}

\label{sec:quant}

We evaluate GEESE on three levels: comparison against baseline methods on within-cohort tasks (Section~\ref{sec:sota_within}), cross-cohort generalization including unified multi-cohort modeling (Section~\ref{sec:cross}), and unified 
multi-cohort phenotyping (Section~\ref{sec:unified}).

\subsubsection{Comparison to SOTA Methods}
\label{sec:sota_within}
We train and evaluate each method independently on each cohort using session-based splits. Baselines fall into three categories: hand-crafted features (raw pose, PCA, s-DANNCE) paired with a linear SVM, learned representations (CEBRA, MOMENT frozen) also paired with a linear SVM, and our end-to-end fine-tuned GEESE. MOMENT (frozen) uses the pretrained encoder without any parameter updates, serving as a direct measure of off-the-shelf foundation model representations. Random guess accuracy reflects class-balanced chance levels. Table~\ref{tab:main_results} summarizes behavior classification and genotype prediction accuracy across all cohorts and methods. 
\begin{table}[!htbp]
\centering
\caption{Downstream task accuracy across methods. Hand-crafted and learned representation methods extract fixed features from 3D pose sequences, on which a linear SVM is trained for each task. GEESE fine-tunes the encoder end-to-end. Genotype prediction for representation-based methods uses direct SVM classification on genotype labels; GEESE uses a two-stage protocol  (behavior training $\rightarrow$ genotype fine-tuning).}
\label{tab:main_results}
\resizebox{\textwidth}{!}{%
\begin{tabular}{llcccccc}
\hline
& & \multicolumn{3}{c}{\textbf{Behavior Classification (Acc.)}} & \multicolumn{3}{c}{\textbf{Genotype Prediction (Acc.)}} \\

\textbf{Method} & \textbf{Representation} & \textbf{CNTNAP2} & \textbf{CHD8} & \textbf{FMR1} & \textbf{CNTNAP2} & \textbf{CHD8} & \textbf{FMR1} \\
\hline
\multicolumn{8}{l}{\textit{Hand-crafted Features + Linear Probe}} \\
Raw pose + SVM          & Joint coordinates          & 57.68\%    & 54.26\%    & 52.70\%    & 53.50\%    & 57.65\%    & 68.67\% \\
PCA + SVM               & PCA components             & 58.60\%    & 53.41\%    & 52.92\%    & 52.28\%    & 56.41\%    & 67.79\% \\
s-DANNCE$^\dagger$~\cite{klibaite2025mapping} & Wavelet + PCA (421-dim) & 75.57\%    & 75.33\%    & 74.38\%    & 57.50\%    & 61.72\%    & 66.45\% \\
\hline
\multicolumn{8}{l}{\textit{Learned Representations + Linear Probe}} \\
CEBRA$^\dagger$~\cite{schneider2023learnable}     & Contrastive embedding     & 64.91\%    & 61.43\%    & 59.84\%    & 42.99\%    & 50.76\%    & 53.43\% \\
MOMENT (frozen)   & Pretrained embedding      & 53.24\%    & 50.92\%    & 49.75\%    & 45.69\%    & 45.86\%    & 44.20\% \\
\hline
\multicolumn{8}{l}{\textit{End-to-End Fine-tuned}} \\
\textbf{GEESE (Ours)}   & \textbf{Fine-tuned embedding} & \textbf{78.56\%} & \textbf{77.99\%} & \textbf{76.30\%} & 58.83\% & 62.82\%    & 67.70\% \\
\hline
\multicolumn{8}{l}{\textit{Reference}} \\
Random Guess                   & --                        & 11.11\%  & 11.11\%  & 11.11\%  & 33.33\%  & 50.00\%  & 50.00\% \\
\hline
\end{tabular}%
}
\end{table}


\noindent\textbf{Behavior Classification.}
We first examine whether training on behavioral labels improves the model's ability to distinguish different behavioral patterns.  
As shown in Table~\ref{tab:main_results}, GEESE outperforms all baselines on behavior classification across all three cohorts (78.56\%, 77.99\%, 76.30\%) without any cohort-specific modification.

s-DANNCE achieves 75.57\%, 75.33\%, and 74.38\% using its 
421-dimensional wavelet+PCA features. GEESE surpasses s-DANNCE on all three cohorts (78.56\% vs.\ 75.57\% on CNTNAP2, 77.99\% vs.\ 75.33\% on CHD8, 76.30\% vs.\ 74.38\% on FMR1) without any hand-crafted features, demonstrating that end-to-end fine-tuning of a pretrained foundation model can outperform carefully designed spectral feature pipelines.

Among learned representations, MOMENT frozen (53.24\%, 50.92\%, 49.75\%) performs comparably to raw pose, confirming that off-the-shelf pretraining does not capture behavioral structure. CEBRA achieves intermediate performance (64.91\%, 61.43\%, 59.84\%). The improvement from frozen to fine-tuned demonstrates that fine-tuning is essential for domain adaptation.

\noindent\textbf{Genotype Prediction.} The model achieves 58.83\% accuracy on CNTNAP2(chance: 33.3\%), 62.82\% on CHD8 (chance: 50.00\%), and 67.70\% on FMR1 
(chance: 50.00\%) through the two-stage protocol (Table~\ref{tab:main_results}). HET animals are most reliably identified (70\%), followed by HOM (47\%) and WT (47\%). The confusion pattern reveals a biologically meaningful gradient: WT is often misclassified as HET (48\%), consistent with the expected gene-dosage effect where loss of one CNTNAP2 copy produces an intermediate phenotype~\cite{cording2025cntnap2}. HOM animals show a more distinct behavioral profile, likely reflecting the more severe social and repetitive behavior abnormalities in complete knockouts.

Note that baseline methods train directly on genotype labels, while GEESE adopts the two-stage protocol described in Section~\ref{sec:training}. Under direct classification, s-DANNCE achieves 57.50\%, 61.72\%, and 66.45\%, consistent with its use of spectral features designed to capture fine-grained kinematic differences. Despite this more constrained protocol, GEESE matches or exceeds s-DANNCE on all three cohorts (58.83\%, 62.82\%, 67.70\%), indicating that the behaviorally-organized manifold not only retains genotype-discriminative information but provides a more effective basis for genotype prediction than features designed purely for kinematic description. Raw pose coordinates also achieve notable accuracy on FMR1 (68.67\%), slightly exceeding GEESE (67.70\%). This suggests that FMR1-associated behavioral differences manifest as detectable postural signatures even without temporal feature extraction, consistent with the pronounced motor phenotype reported in Fragile X models~\cite{chen2022early}.

We compared GEESE against ETSformer~\cite{woo2022etsformer} and DLinear~\cite{zeng2023transformers} as alternative backbones; MOMENT-Large consistently outperformed both in frozen and fine-tuned settings (Appendix~\ref{tab:ablation}), confirming that large-scale pretraining provides a stronger initialization for behavioral classification.

\subsubsection{Cross Cohort Generalization}
\label{sec:cross}

We evaluate whether representations transfer across genetic backgrounds by applying source-trained models to unseen target cohorts, comparing against s-DANNCE as the strongest baseline.

\noindent\textbf{Cross-cohort Behavior Transfer.}
Table~\ref{tab:cross_cohort} reports behavior classification accuracy when a model trained on one cohort is applied directly to another without retraining. For s-DANNCE, this involves fitting PCA on the source cohort and computing features for the target cohort using the source PCA basis, then applying the source-trained classifier. For GEESE, the source-trained encoder and classification head are applied directly to target cohort data.

GEESE consistently outperforms s-DANNCE in both within-cohort (diagonal) and cross-cohort (off-diagonal) settings. Notably, GEESE models trained on CNTNAP2 achieve 76.4\% on CHD8 and 75.2\% on FMR1, retaining within-cohort performance. The relatively small drop from diagonal to off-diagonal for both methods suggests that the nine behavioral categories share consistent kinematic signatures across genetic backgrounds. However, GEESE's larger advantage in the cross-cohort setting indicates that learned representations capture more transferable behavioral structure than hand-crafted spectral features.

\begin{table}[!htbp]
\centering
\caption{Cross-cohort generalization. Behavior classification uses 
zero-shot transfer; genotype classification trains an MLP on 30\% of target labels. Rows indicate source cohort; columns indicate target. Diagonal entries (shaded) are within-cohort results from Table~\ref{tab:main_results}.}
\label{tab:cross_cohort}
\resizebox{0.85\textwidth}{!}{%
\begin{tabular}{llccc|ccc}
\hline
& & \multicolumn{3}{c|}{\textbf{Behavior (zero-shot)}} & \multicolumn{3}{c}{\textbf{Genotype (30\% labels)}} \\
& \textbf{Source $\backslash$ Target} & \textbf{CNTNAP2} & \textbf{CHD8} & \textbf{FMR1} & \textbf{CNTNAP2} & \textbf{CHD8} & \textbf{FMR1} \\
\hline
\multirow{3}{*}{\textbf{s-DANNCE$^\dagger$}}
& CNTNAP2  & \cellcolor{gray!20}75.57 & 72.33 & 73.04 & \cellcolor{gray!20}57.50 & 57.00 & 56.31 \\
& CHD8     & 71.26 & \cellcolor{gray!20}75.33 & 71.38 & 59.81 & \cellcolor{gray!20}61.72 & 59.56 \\
& FMR1     & 72.71 & 70.64 & \cellcolor{gray!20}74.38 & 62.94 & 63.56 & \cellcolor{gray!20}66.45 \\
\hline
\multirow{3}{*}{\textbf{GEESE}}
& CNTNAP2  & \cellcolor{gray!20}78.56 & 76.35 & 75.22 & \cellcolor{gray!20}58.83 & 57.30 & 57.21 \\
& CHD8     & 72.41 & \cellcolor{gray!20}77.99 & 71.35 & 60.83 & \cellcolor{gray!20}62.82 & 61.35 \\
& FMR1     & 75.10 & 73.95 & \cellcolor{gray!20}76.30 & 61.74 & 61.36 & \cellcolor{gray!20}67.70 \\
\hline
\end{tabular}%
}
\end{table}

\noindent\textbf{Cross-cohort Genotype Classification.}
While cross-cohort behavior transfer applies a source classifier directly to target data, cross-cohort genotype evaluation requires a different protocol: the genes differ across cohorts, so a genotype classifier trained on CNTNAP2 (WT/HET/HOM) cannot be applied to CHD8 (WT/HET). Instead, we evaluate whether a source cohort's representation supports genotype classification on a target cohort when given limited target labels. Specifically, we use source-trained feature extractors (GEESE encoder or s-DANNCE PCA basis) to compute target cohort representations, then train an MLP classifier on 30\% of the target cohort's genotype labels and evaluate on the remaining data. Table~\ref{tab:cross_cohort} reports the results.

Both methods show comparable performance with off-diagonal entries close to diagonal values, indicating that learned representations preserve genotype-discriminative information across genetic backgrounds.

Notably, s-DANNCE represents a strong baseline derived from domain-expert-designed spectral features~\cite{klibaite2025mapping}; GEESE's comparable or superior performance across all source-target pairs demonstrates that learned representations match expert-crafted features in cross-cohort transferability.

\subsubsection{GEESE as A Unified Model for Multi-cohort Phenotyping}
\label{sec:unified}

To evaluate whether a single model can perform phenotyping across all genetic backgrounds simultaneously, we trained a single model on combined data from all three cohorts and evaluated on two tasks: overall behavior classification and unified 7-class genotype identification (CNTNAP2-WT/HET/HOM, CHD8-WT/HET, FMR1-WT/HET). Table~\ref{tab:joint_training} reports the results. GEESE achieves 74.17\% behavior accuracy, comparable to the average of cohort-specific models, confirming that a single encoder can represent behavioral structure across diverse genetic backgrounds without cohort-specific tuning. On the 7-class genotype task (chance: 14.29\%), GEESE achieves 59.47\% compared to s-DANNCE's 57.73\%, demonstrating that both methods can distinguish not only genotype dosage within a cohort but also genetic background across cohorts from behavioral data alone.

\begin{table}[ht!]
\centering
\caption{Joint training across all cohorts. A single model is trained on combined data from all three cohorts. Behavior classification accuracy is reported on the combined test set. Genotype classification is a 7-class task (CNTNAP2-WT/HET/HOM, CHD8-WT/HET, FMR1-WT/HET).}
\label{tab:joint_training}
\begin{tabular}{lcc}
\hline
\textbf{Method} & \textbf{Behavior (overall)} & \textbf{Genotype (7-class)} \\
\hline
s-DANNCE  & 71.55\% & 57.73\% \\
GEESE     & \textbf{74.17}\% & \textbf{59.47}\% \\
\hline
\end{tabular}
\end{table}

\subsubsection{Ablation Study}
\label{sec:ablation}

Table~\ref{tab:ablation} compares different foundation model configurations on behavior classification to isolate the contributions of model architecture and fine-tuning.

In the frozen setting, all three models perform comparably, confirming that pretrained representations alone do not capture rodent behavioral structure. Fine-tuning reveals clear differences: ETSformer plateaus below 70\%, DLinear shows inconsistent gains, while GEESE (MOMENT-Large) achieves the highest accuracy across all cohorts (78.56\%, 77.99\%, 76.30\%). The gap between frozen and fine-tuned MOMENT indicates that the value of the foundation model lies in its learned inductive biases that facilitate efficient domain adaptation through fine-tuning.

\begin{table}[!htbp]
\centering
\caption{Ablation study: foundation model comparison}
\label{tab:ablation}
\begin{tabular}{llccc}
\hline
\textbf{Model} & \textbf{Configuration} & \textbf{CNTNAP2} & \textbf{CHD8} & \textbf{FMR1} \\
\hline
\multicolumn{5}{l}{\textit{Time Series Foundation Models}} \\
MOMENT-Large & Frozen & 54.45\% & 53.16\% & 56.42\% \\
ETSformer~\cite{woo2022etsformer} & Frozen & 57.80\% & 53.27\% & 56.22\% \\
DLinear~\cite{zeng2023transformers} & Frozen & 58.95\% & 55.56\% & 53.51\% \\
\hline
\multicolumn{5}{l}{\textit{Fine-tuned Configurations}} \\
ETSformer~\cite{woo2022etsformer} & Full fine-tune & 69.55\% & 66.07\% & 67.81\% \\
DLinear & Full fine-tune & 62.93\% & 67.74\% & 65.52\% \\
\textbf{GEESE (Ours)} & Full fine-tune & \textbf{78.56\%} & \textbf{77.99\%} & \textbf{76.30\%} \\
\hline

\end{tabular}
\end{table}

\subsection{Qualitative Evaluation}

\subsubsection{Learned Representations Reveal Interpretable Behavioral Structure} 
\label{sec:behav_clusters}

To examine this, we applied K-Means clustering~\cite{hartigan1979algorithm} (k=9) to the learned embeddings without using any labels. 

\begin{figure}[!htbp]
\centering
\begin{subfigure}{0.4\textwidth}
    \centering
    \includegraphics[width=\textwidth, trim=0 0 500 0, clip]{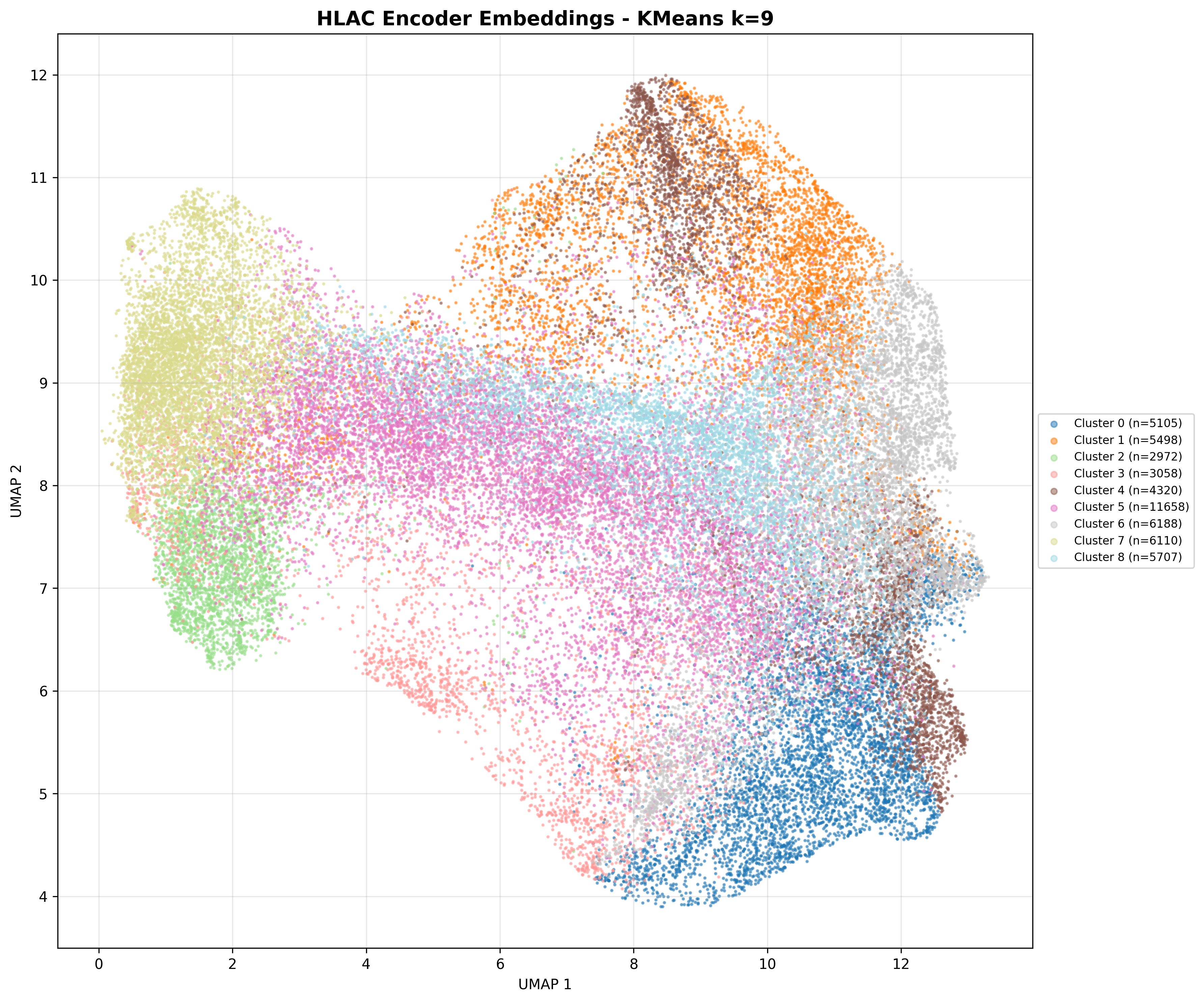}
    \subcaption{Behavioral clusters}
    \label{fig:behavior_clusters}
\end{subfigure}
\hfill
\begin{subfigure}{0.45\textwidth}
    \centering
    \includegraphics[width=\textwidth]{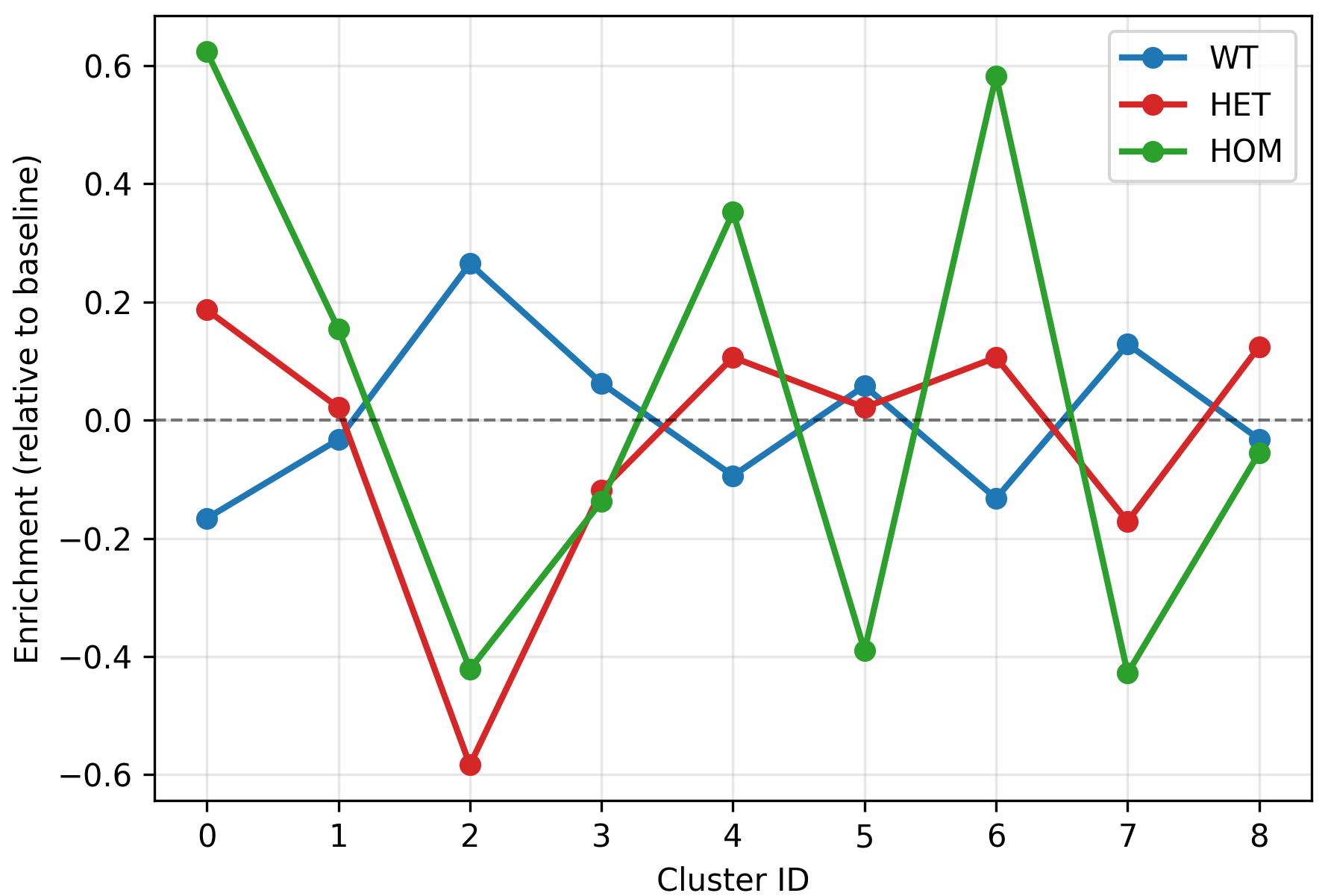}
    \subcaption{Genotype enrichment}
    \label{fig:geno_enrichment}
\end{subfigure}

\vspace{4pt}

\begin{subfigure}{0.15\textwidth}
    \centering
    \includegraphics[width=\textwidth]{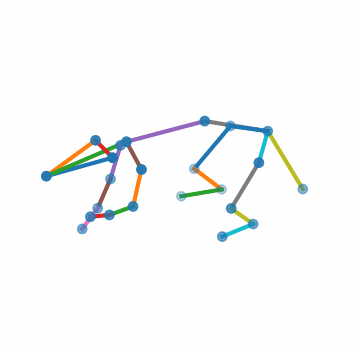}
    \subcaption{Idle}
    \label{fig:skel_idle}
\end{subfigure}
\hfill
\begin{subfigure}{0.15\textwidth}
    \centering
    \includegraphics[width=\textwidth]{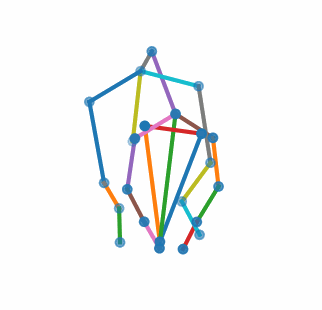}
    \subcaption{Grooming}
    \label{fig:skel_grooming}
\end{subfigure}
\hfill
\begin{subfigure}{0.15\textwidth}
    \centering
    \includegraphics[width=\textwidth]{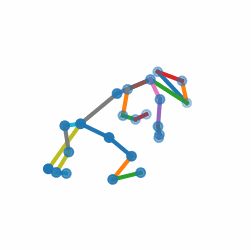}
    \subcaption{Locomotion}
    \label{fig:skel_locomotion}
\end{subfigure}
\hfill
\begin{subfigure}{0.15\textwidth}
    \centering
    \includegraphics[width=\textwidth]{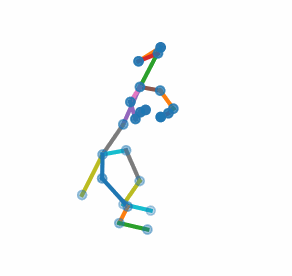}
    \subcaption{Rearing}
    \label{fig:skel_rearing}
\end{subfigure}
\hfill
\begin{subfigure}{0.15\textwidth}
    \centering
    \includegraphics[width=\textwidth]{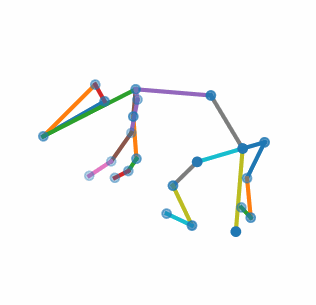}
    \subcaption{Head-turning}
    \label{fig:skel_headturn}
\end{subfigure}
\caption{Learned behavioral representations. (a) Behavioral clusters via K-Means (k=9), visualized with UMAP. (b--f) Representative skeleton sequences. (g) Genotype enrichment per cluster, showing differential distribution of WT, HET, and HOM animals.}
\label{fig:cluster_and_enrichment}
\end{figure}

For visualization, we project embeddings to two dimensions using UMAP~\cite{mcinnes2018umap} with cosine distance.  
The grouping revealed nine distinct clusters corresponding to different movement patterns (Figure~\ref{fig:behavior_clusters}). Visual inspection of representative skeleton sequences (Figure~\ref{fig:skel_idle}~\ref{fig:skel_grooming}~\ref{fig:skel_locomotion}~\ref{fig:skel_rearing}~\ref{fig:skel_headturn}) shows that each cluster captures characteristic behaviors: Cluster 0 contains primarily idle and forward looking postures; Cluster 2 and 8 are characterized by head-turning and lateral head-turning; Cluster 4 shows face-grooming; Cluster 5 reflects rearing and forward-locomotion; Cluster 7 captures rearing and ground-sniffing. We further compared cluster assignments with human-labeled behavioral categories to quantify how well these groups correspond to expert annotations. Behaviors with distinctive motion signatures (e.g., fast locomotion, grooming) map cleanly to individual clusters, while kinematically similar behaviors show more overlap.


\subsubsection{Genotype Signatures Emerge in Learned Behavioral Representations}

The quantitative results in Section~\ref{sec:quant} establish that GEESE can classify both behavior and genotype. Here we examine how genotype information is organized within the learned representation, and how fine-tuning transforms the manifold from a behavior-only space into one that jointly encodes behavior and genotype.

\noindent\textbf{Genotype Signatures in Behavior-trained Representations.}
We first asked whether the behavior-trained embedding space already contains genotype information, before any genotype-specific training. We examined how WT, HET, and HOM animals distribute across the behavioral clusters identified in Section~\ref{sec:behav_clusters}.

Genotype distributions vary substantially across clusters (Figure~\ref{fig:geno_enrichment}), with HOM enriched in Clusters 0, 4, and 6. These clusters correspond to idle postures and repetitive grooming movements, consistent with the reduced exploration and increased stereotypy reported in CNTNAP2 knockout models~\cite{valeeva2024role}. This indicates that genotype-associated behavioral patterns are already present in the embedding space after behavioral category training alone, suggesting that genotype and behavior information naturally coexist in the learned representation.


\noindent\textbf{Fine-tuning Transforms the Manifold into a Genotype-aware Space.}
Although genotype information exists in the behavior-trained manifold, the model was not explicitly trained to detect it. We fine-tuned the encoder on genotype labels with a reduced learning rate, aiming to preserve learned behavioral structure while gaining sensitivity to genotype.

\begin{figure}[!htbp]
\centering
\begin{subfigure}{0.4\textwidth}
    \includegraphics[width=\textwidth]{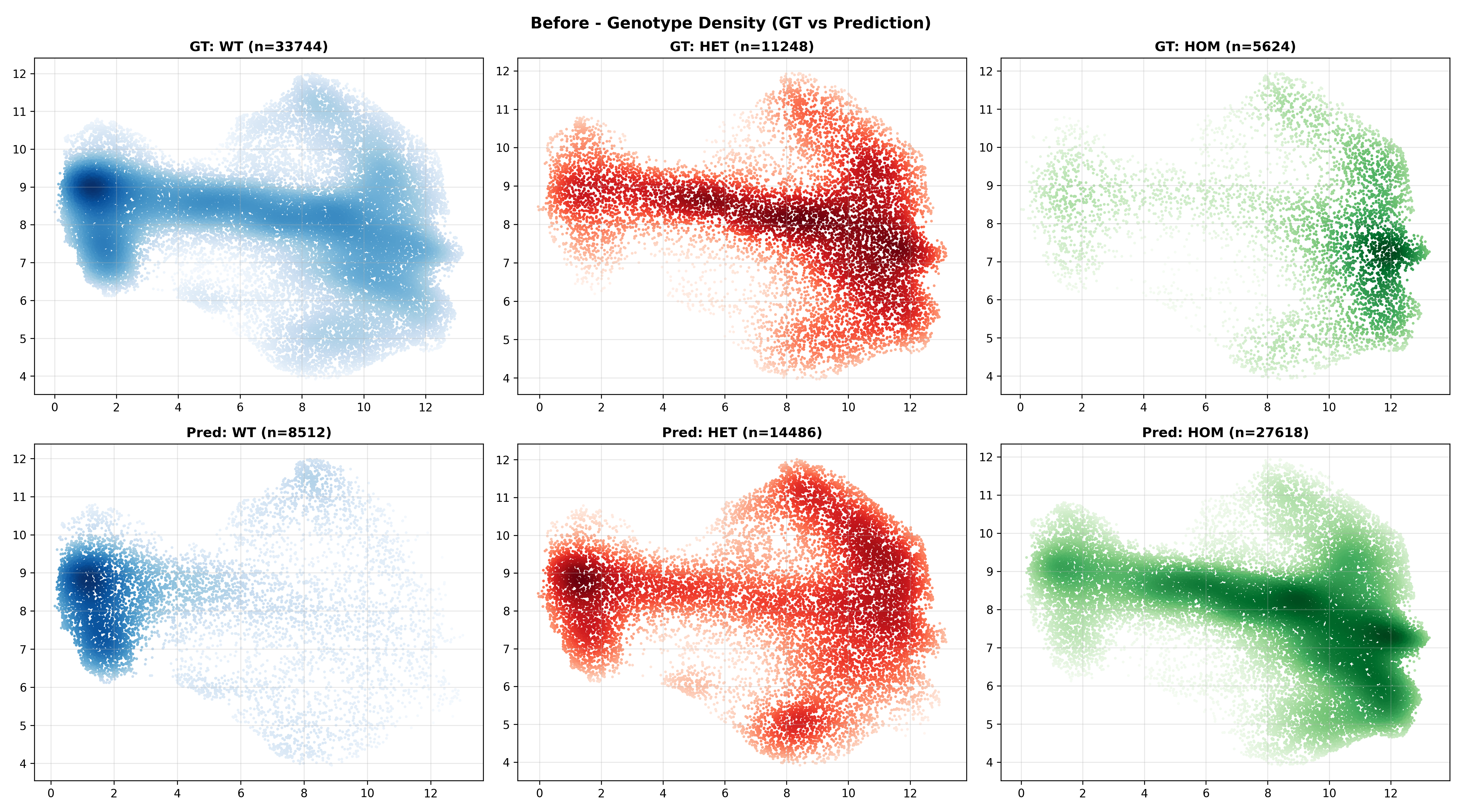}
    \caption{Before genotype finetuning}
    \label{fig:before_genoft}
\end{subfigure}
\begin{subfigure}{0.4\textwidth}
    \includegraphics[width=\textwidth]{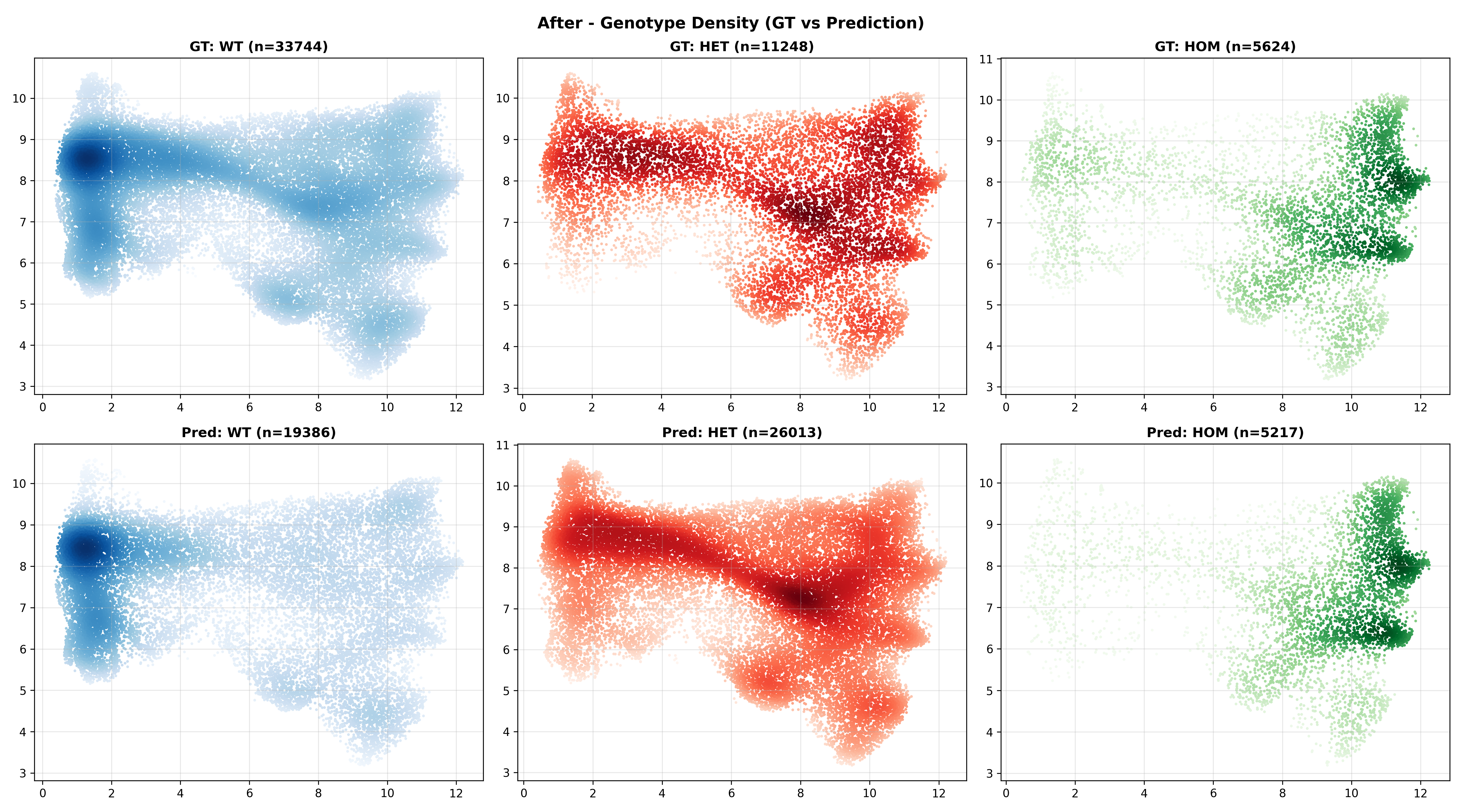}
    \caption{After genotype finetuning}
    \label{fig:after_genoft}
\end{subfigure}

\caption{Genotype-behavior association learning. (a) Before and (b) after genotype fine-tuning. Top row: ground truth genotype distributions; bottom row: predicted distributions. Columns show WT (blue), HET (red), and HOM (green). After fine-tuning, predicted distributions converge toward ground truth, with HOM occupying a distinct region.}
\label{fig:genotype_behavior}
\end{figure}

Before fine-tuning, predicted genotype distributions show poor correspondence with ground truth (Figure~\ref{fig:before_genoft}). After fine-tuning, predicted distributions align more closely with ground truth: WT predictions expand toward the center, better matching the true distribution, HET spans the central area, and HOM forms a distinct cluster in the lower-right corner (Figure~\ref{fig:after_genoft}). The particularly clear separation of HOM indicates that complete knockouts exhibit more distinctive behavioral signatures. Critically, this genotype separation emerges within the existing behavioral structure. The behavioral clusters remain intact while gaining genotype sensitivity. The resulting representations encode both behavior type and genotype dosage in a shared space, enabling the model to capture not only what an animal is doing but how its genetic background modulates that behavior.

To quantify this transformation, we computed mean squared error (MSE) between predicted and ground truth genotype enrichment per behavior class. After fine-tuning, MSE decreases from 0.465 to 0.287 (38\% reduction), with particularly notable gains in Idle, Locomotion, and Fast Locomotion, where CNTNAP2 models have shown altered activity levels and repetitive patterns~\cite{valeeva2024role}. The behavior-specific nature of this improvement confirms that the model captures genotype differences as they manifest within individual behavioral contexts, rather than learning a global genotype signal divorced from behavior.

\subsection{An Interactive and Intelligent Toolbox: from Manifold to Real-time Phenotyping}

\begin{figure}[!htbp]
\centering
\includegraphics[width=\textwidth]{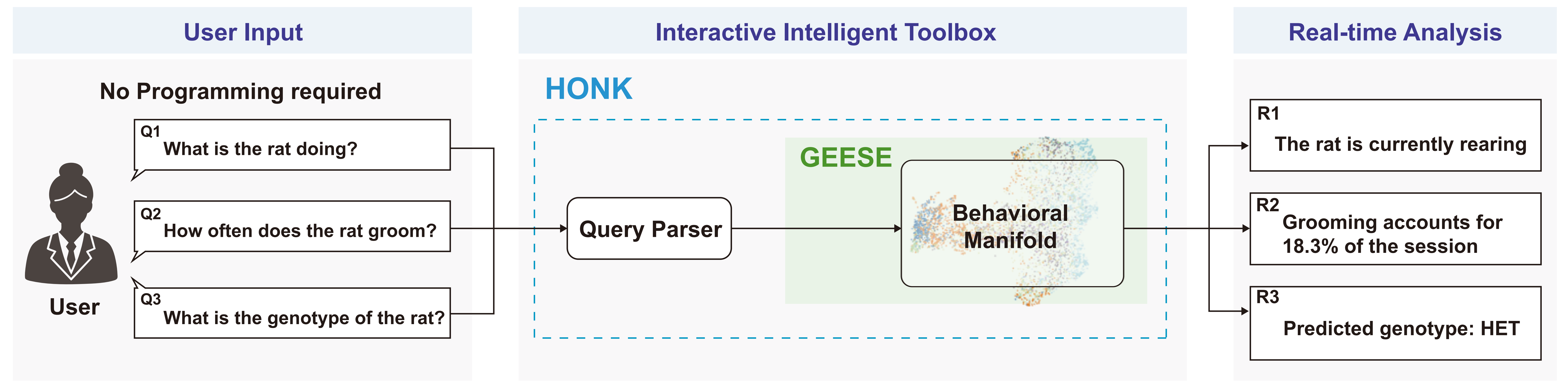}
\caption{Concept diagram of HONK, the interactive analysis agent built on the GEESE pipeline. Users pose natural language queries that are routed to the pipeline, which encodes 3D pose data into a behavioral manifold for behavior classification, genotype prediction, and interactive visualization. The system provides real-time analysis without requiring programming expertise or local computational resources.}
\label{fig:honk}
\end{figure}

To make these methods accessible to users without programming expertise, we developed HONK (Hands-On Natural-language Knowledgebase), an interactive analysis agent built on the GEESE pipeline for behavioral phenotyping from pose dynamics (Figure~\ref{fig:honk}).

\noindent\textbf{Functionality and Accessibility. }
Given a raw pose sequence, HONK processes it through the all-cohort model and returns predicted behavior distributions, genotype probabilities (including 7-class identification), temporal behavior sequences, and manifold visualizations. Users interact through natural language queries (e.g., "What is the predicted genotype?"). The platform prioritizes interpretability: users can trace how individual windows map onto the manifold and export results for downstream analysis. HONK requires no local installation or computational resources. A web-based deployment of HONK will be available soon.

\section{Conclusion}

We present GEESE, an end-to-end framework that learns behavioral representations directly from 3D pose dynamics using a pretrained foundation model, eliminating hand-crafted feature engineering entirely. Across three autism-associated genetic models, GEESE surpasses the leading hand-crafted baseline in both behavior classification and genotype prediction. A single model trained on all cohorts identifies genetic background and genotype from movement patterns alone, enabling unified phenotyping without prior knowledge of the experimental cohort. The accompanying tool HONK makes this pipeline accessible to researchers through natural language interaction, providing a step toward automated, reproducible behavioral assessment in genetic disease models.

\noindent\textbf{Limitations and Future Directions.}
The current framework operates on single-animal 3D pose data, providing a controlled setting for isolating genotype-behavior associations; extending to multi-animal social interactions and integrating additional modalities such as facial features are natural next steps. The reliance on 3D pose estimation is mitigated by the architecture's agnosticism to spatial dimensionality, substituting 2D pose inputs from tools such as DeepLabCut~\cite{mathis2018deeplabcut} or SLEAP~\cite{pereira2022sleap} requires only a change in input channel dimension. Incorporating explicit spatial relationships between joints into the encoder is a promising direction for capturing coordinated movement patterns that may carry additional phenotypic information.Validation on independent datasets from other laboratories would further establish the generalizability of the learned representations.

\renewcommand{\refname}{\centering References}

\bibliography{main}

\appendix

\end{document}